\documentclass[preprint, sort&compress]{elsarticle}

\usepackage{amsmath}
\usepackage{amssymb}
\usepackage{graphicx}
\usepackage{amsfonts}
\usepackage{rotating}

\usepackage{bm}
\usepackage{subfigure}
\usepackage{booktabs}
\usepackage{multirow}
\usepackage{diagbox}

\newdefinition{Def}{Definition}
\newtheorem{prop}{Proposition}
\newdefinition{remark}{Remark}

\begin{document}

\begin{frontmatter}

\title{A Degree of Flowability for Virtual Tubes}

\author[1]{Quan Quan}
\author[1]{Shuhan Huang}
\author[1]{Kai-Yuan Cai}

\affiliation[1]{organization={School of Automation
		Science and Electrical Engineering, Beihang University},
	addressline={37 Xueyuan Road, Haidian District}, 
	city={Beijing},
	postcode={100191}, 
	country={P.R. China}}

\begin{abstract}

With the rapid development of robotics swarm technology, there are more tasks that require the swarm to pass through complicated environments safely and efficiently. Virtual tube technology is a novel way to achieve this goal. Virtual tubes are free spaces connecting two places that provide safety boundaries and direction of motion for swarm robotics. How to determine the design quality of a virtual tube is a fundamental problem. For such a purpose, this paper presents a degree of flowability (DOF) for two-dimensional virtual tubes according to a minimum energy principle. After that, methods to calculate DOF are proposed with a feasibility analysis. Simulations of swarm robotics in different kinds of two-dimensional virtual tubes are performed to demonstrate the effectiveness of the proposed method of calculating DOF.

\end{abstract}


\end{frontmatter}

\section{Introduction}
In recent years, advances in robotic swarm control have brought various applications, such as environment surveillance and mapping, air transportation and delivery, searching and rescuing, and so on \cite{schranz2020swarm}. These applications raise a high demand for the robotic swarm to operate in cluttered environments \cite{chung2018survey}. There are several types of control methods for the robotic swarm to pass through a complicated environment \cite{gao2023survey}, including leader-follower formation control, multi-robot trajectory planning, control-based methods, and virtual tube technology. The formation control keeps the robotic swarm maintained in a particular geometric structure, which usually needs to identify neighboring robots’ IDs \cite{Henrik2021comparative} and lacks consideration for traffic efficiency. The multi-robot trajectory planning methods focus on continuous trajectories for robotic swarms using centralized or distributed algorithms \cite{xia2021multi}. It relies on direct communication, and is suitable for an individual robot or small teams rather than large-scale swarms. The control-based methods provide instant control instructions for each robot \cite{yuri2022through}. It fits the control for a large swarm without ID identification and direct communication but lacks consideration for efficiency and may cause deadlocks.

Virtual tube technology combines the advantages of both trajectory planning and the control-based method \cite{quan2023distributed}. Instead of planning a trajectory for each robot, the method first plans a two-dimensional or three-dimensional tube, called the \emph{virtual tube} in which there are no obstacles \cite{mao2023optimal}. Then, the control-based method drives the robotic swarm passing through the virtual tube and keeping the swarm within the virtual tube while avoiding collision among robots \cite{quan2023distributed}.

Further questions are raised on how to decide whether a virtual tube is easy for the swarm to pass through. The assessment of trafficability and flowability has been researched in many other fields of science and technology. In geoscience field, trafficability is thought related to geographical and geological data. For instance, the study \cite{he2023assessment} makes use of the Analytic Hierarchy Process-Weighted Information Content (AHP-WIC) method, the clustering method and the watershed method to process those data and assess off-road trafficability. In the fields of powder mechanics, flowability is used to describe the flowing capacity of powder \cite{bell2001solids}, and a characterization of flowability is the flow index, which is the ratio of consolidation stress to unconfined yield strength \cite{ashish2019effect}. In pipeline engineering, the energy loss of fluid flowing through a tube is often considered. Pipe is a type of tube, and the major energy loss on pipe is due to friction while the minor energy loss is caused by the change of section and barriers in the tube \cite{sunaris2019analysis}. The friction coefficient is related to the Reynolds Number and roughness of pipe wall \cite{lahiouel2015evaluation}.

To the best of the authors’ knowledge, no research has considered the fundamental problem about the measure of traffic capacity for a virtual tube. Navigating a robotic swarm through a narrow and curved corridor or towards a common destination may result in congestion and a reduction in the efficiency of passage \cite{sharma2022priority} \cite{yuri2023congestion}. Consequently, the assessment of the trafficability of virtual tubes is a crucial aspect of path planning and control for the robotic swarm. Our aim is to evaluate the difficulty of controlling a robotic swarm to pass through, which is different from the concepts of trafficability and flowability mentioned before. But we can borrow some of their ideas to construct our measure to assess the traffic capacity for virtual tubes. The robotic swarm moving in the virtual tube has a variable shape and flows like the solid particles in a two-phase flow. To simplify the model, we ignore the size of the robots and treat the robotic swarm as a continuous fluid flow, enabling a straightforward calculation. In this situation, the process of the swarm passing through the virtual tube can be described as the process of a segment of fluid flowing through a tube. Although the robotic swarm is much different from fluid flow, there still exists friction between the swarm and the environment. Therefore, we use the term "flowability" to describe the trafficability of a virtual tube, and add friction to key factors of flowability.

The energy required for a process has been identified as a key metric in a number of different fields, in addition to the pipeline engineering previously discussed. In the field of control theory, a degree of controllability (DOC) is used to describe the extent to which a system can be controlled \cite{kalman1963controllability}. A widely-used measure of DOC is based on the minimum control input energy to change states of the dynamic system. Study \cite{muller1972analysis} proposed three measures of DOC based on Controllability Gramian matrix, which is related to the minimum energy input to control a system from an initial state to zero state within a finite time. In the field of slurry pipeline transport, specific energy consumption is an important measure of transport efficiency \cite{hashemi2014specific}. The minimum specific energy consumption has been analyzed to obtain the optimal transport concentration \cite{li2022analysis}. These findings inspire us to use the minimum energy input as a measure of the degree of flowability (DOF) for virtual tubes defined in our previous work \cite{mao2022making} in this paper. Methods to calculate DOF are proposed with a feasibility analysis. The main contributions of this paper are:
\begin{itemize}
	\item DOF for two-dimensional virtual tubes is defined.
	\item The method to calculate the DOF for a two-dimensional virtual tube is presented, and the feasibility of the method is also analyzed. The effectiveness of the proposed method is validated by the simulations.
\end{itemize}

\section{Problem Formulation}

\subsection{Two-Dimensional Virtual Tube}

According to our previous work \cite{mao2022making}, a rigorous definition of the virtual tube in the two-dimensional Euclidean space is presented.

\begin{Def}
	\label{def: 2d virtual tube}
	A virtual tube in two-dimensional Euclidean space is a subset $\mathcal{T}$ of $\mathbb{R}^2$ parameterized as
	\begin{equation}
		{\mathcal{T}} \left( s, \theta, \lambda \right) = {\bm{\gamma}} \left( s \right) +
		\lambda r \left( s, \theta \right) \mathbf{n} \left( s \right) \cos \theta,
		\label{equ: 2d virtual tube}
	\end{equation}
	where $s \in {\mathcal{D}}_s= \left[ {{s_0},{s_f}} \right] \subset \mathbb{R}$ determines a certain point along the generating curve; $\theta \in {\mathcal{D}}_{\theta} = \left\{ 0, \pi \right\}$; $\lambda \in {\mathcal{D}}_{\lambda} = \left[0, 1\right]$; ${\mathcal{T}} \left( s, \theta, \lambda \right)$ is a position of a point in the virtual tube, defined by parameters $s$, $\theta$ and $\lambda$. The second order differentiable curve ${\bm{\gamma }}\left( s \right) \in \mathbb{R}^2$ is the \emph{generating curve}; $r \left( {s,\theta } \right) \in \mathbb{R}$ called \emph{radius} is continuous and differentiable with respect to $s$ and $\theta$, and satisfies $r \left( {s,\theta } \right) > 0$ for any $s \in \mathcal{D}_s$, $\theta \in \mathcal{D}_{\theta}$, whose direction is determined by $\cos \theta$. Given $s_1 \in {\mathcal{D}}_s$, $\mathbf{t}\left( s_1 \right) = \frac{{{\bm{\dot \gamma }}\left( s_1 \right)}}{{\Vert {{\bm{\dot \gamma }}\left( s_1 \right)} \Vert}}$ is the \emph{unit tangent vector} of the generating curve at point $\bm{\gamma} \left( s_1 \right)$; $\mathbf{n}\left( s_1 \right)$ is the \emph{normal vector} of the generating curve at point $\bm{\gamma}\left(s_1 \right)$; parameter $\lambda$ is the proportion of the distance between a certain point ${\mathcal{T}}\left( s_1, \theta, \lambda \right) $ and the point ${\bm{\gamma}} \left( s_1 \right) $ to the radius $r\left( s_1,\theta \right) $. A \emph{virtual tube cross section} at point ${\bm{\gamma}} \left(s_1\right)$ is defined by $\mathcal{T}_{s_1}\left( {\theta, \lambda } \right) = \mathcal{T}\left( {s_1,\theta, \lambda} \right)$; the width of the virtual tube is the length of the cross section, denoted by $l \left(s_1\right) = r \left( s_1, 0 \right) + r \left( s_1, \pi \right)$. The \emph{virtual tube boundary} is defined by $\mathcal{T}_1\left( {s,\theta } \right) = {\mathcal{T}}\left( {s,\theta ,1 } \right) = {\bm{\gamma}} \left( s \right) + r \left( s, \theta \right) \mathbf{n} \left( s \right) \cos \theta$, where $r \left(s, \theta \right)$ varies with parameters $s$ and $\theta$ to decide the shape of the virtual tube.
\end{Def}

\begin{figure}[htbp]
	\centering
	\includegraphics[width=0.7\linewidth]{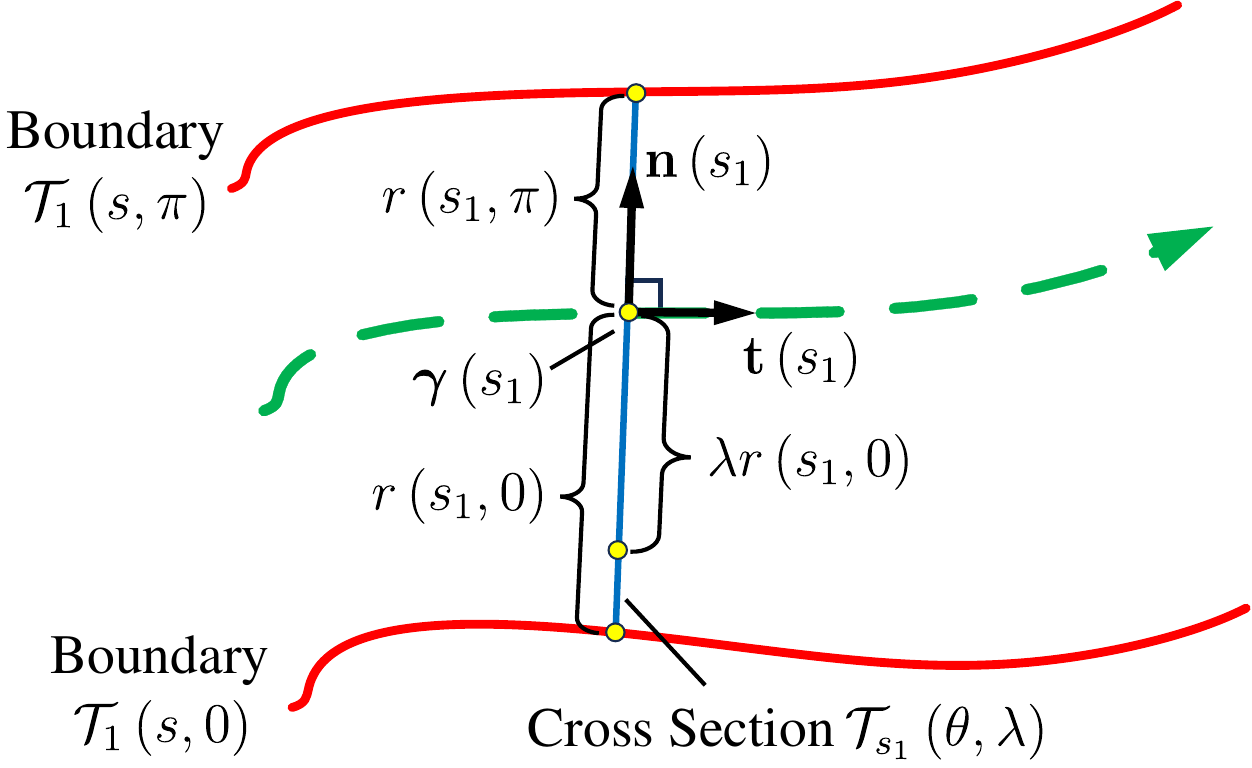}  
	\caption{A two-dimensional virtual tube \cite{quan2023distributed}}
	\label{fig: 2d virtual tube}
\end{figure}

A virtual tube ${\mathcal{T}}$ is generated by keeping the cross section $\mathcal{T}_{s}\left( {\theta,\lambda } \right)$ always orthogonal to the
tangent vector $\mathbf{t} \left( s \right) $ of the generating curve $\bm{\gamma}$, denoted by 
\begin{equation}
	{\mathcal{T}} = \bigcup\limits_{s \in \mathcal{D}_s} \mathcal{T}_{s}\left( {\theta,\lambda } \right).
\end{equation}

Furthermore, the velocity of the cross section is defined.

\begin{Def}
	\label{def: velocity of cross section}
	For virtual tube ${\mathcal{T}} \left( s, \theta, \lambda \right)$, given $s_1 \in {\mathcal{D}}_s$, the \emph{velocity of the cross section} at $s_{1}$ is the velocity of point ${\bm{\gamma}} \left( s_{1}\right)$ on the generating curve. The magnitude of the velocity is the \emph{speed of the cross section} at point ${\bm{\gamma}}\left( s_{1}\right)$.
\end{Def}

\subsection{Degree of Flowability}

As mentioned above, a virtual tube is generated by a length-varying cross section moving perpendicularly along a generating curve. The energy consumed from moving the cross section is considered when measuring DOF of a virtual tube, which is consistent with the definition of DOC. Besides, a wider virtual tube has a larger capacity than a narrow one, and robots in the swarm may spend less time and energy avoiding each other in a wider virtual tube. Therefore, the width of the virtual tube is also taken into account. In order to calculate the energy consumed, the cross section is set with mass. The gravitational potential is ignored because it only depends on the entrance and exit of a virtual tube. On the other hand, the work done by resistance is considered. Based on the description above, the definition of DOF is derived.

\begin{Def}
	\label{def: dof}
	The \emph{degree of flowability} (DOF) of a virtual tube ${\mathcal{T}}\left( {s, \theta, \lambda} \right)$ is a function of the average width and the minimum input energy to move the cross section $\mathcal{T}_{s}\left( {\theta, \lambda} \right)$ along the generating curve ${\bm{\gamma}}\left( s \right)$ from starting point ${\bm{\gamma}}\left( s_0 \right)$ to finishing point ${\bm{\gamma}}\left( s_f \right)$, expressed as
	\begin{equation}
		\label{equ: dof expression}
		DOF = \frac{\overline{l}}{E},
	\end{equation}
	where $E$ is the minimum energy input, $\overline{l}$ is the average width of the virtual tube defined as
	\begin{equation}
		\overline{l} = \frac{1}{s_{\rm f} - s_0} \cdot \int_{s_0}^{s_{\rm f}} l \left(s\right) {\rm d} s.
		\label{equ: average width}
	\end{equation}
\end{Def}

\begin{remark}
	In the classic definition of DOC, controllability Gramian matrix is used to calculate the optimal control law based on the minimum-energy principle \cite{kalman1963controllability} \cite{muller1972analysis}. Here the DOF of a virtual tube is a measure that relates the energy cost for objects to pass through it. The energy cost is positively associated with the length of the generating curve as well as the change of the cross section. With reference to DOC, the DOF is set inversely proportional to the minimum energy input. It is clearly unreasonable that straight virtual tubes with different widths have the same DOF value when using only the energy metric. Therefore, the average width is used in the measure of DOF as well. A larger width of the virtual tube will decrease the time spent in avoiding collision and have more freedom for traffic, just like a wider highway allows more traffic flow and reduces possibility of congestion \cite{krishnamurthy2014effect}. Accordingly, the DOF is commensurate with the mean length of the virtual tube.
	\label{remark: DOC and DOF}
\end{remark}

\begin{figure*}[!htbp]
	\centering
	\subfigure[Tube 1: a straight tube with a straight generating curve and a constant radius]{
		\centering
		\includegraphics[width=0.286\linewidth]{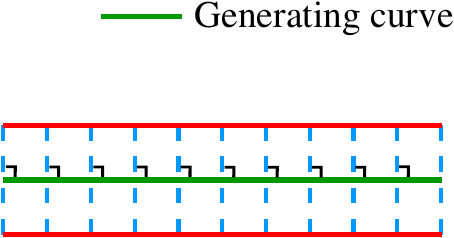}
		\label{subfig: Tube 1}}
	\quad
	\subfigure[Tube 2: a curving tube with a curved generating curve and a constant radius]{
		\centering
		\includegraphics[width=0.286\linewidth]{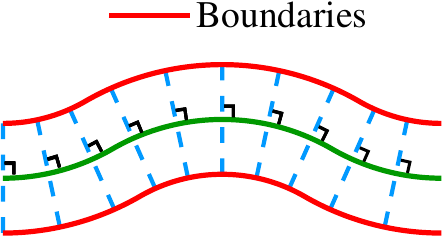}
		\label{subfig: Tube 2}}
	\quad
	\subfigure[Tube 3: a straight tube with a curved generating curve]{
		\centering
		\includegraphics[width=0.286\linewidth]{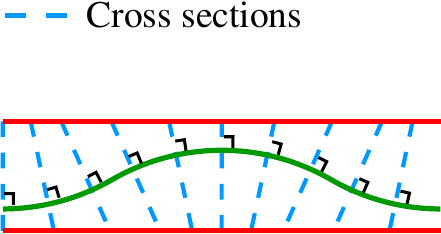}
		\label{subfig: Tube 3}}
	\caption{Three kinds of 2-D virtual tubes}
	\label{fig: three examples}
\end{figure*}

This paper aims to provide a rationale for the concerns raised about energy costs in the definition. To this end, a simple analysis of the different kinds of virtual tubes is presented. Three typical two-dimensional virtual tubes are shown in Fig.~\ref{fig: three examples}.

\begin{itemize}
	\item \textbf{Tube 1.} A straight virtual tube whose generating curve is straight and the cross section has a constant length. 
	
	\item \textbf{Tube 2.} A curve virtual tube whose generating curve is a continuous curve and the cross section has a constant length. 
	
	\item \textbf{Tube 3.} A straight virtual tube whose generating curve is a continuous curve, with a length-varying cross section.
\end{itemize}

\begin{remark}
	To make better comparisons, assume that the three virtual tubes have the same starting cross section and finishing cross section, and their cross sections possess a same mass and move at a same speed.
\end{remark}

Intuitively, \textit{Tube 1} has a larger DOF than \textit{Tube 2} and \textit{Tube 3}. Preliminary analysis of energy change when the cross section passes through two virtual tubes is implemented.

\begin{itemize}
	\item Cross section of \textit{Tube 1} moves the shortest distance, resulting in the least energy loss from resistance. 
	
	\item Cross section of \textit{Tube 2} moves a longer distance than \textit{Tube 1}, and so causes a larger energy input for resistance. 
	
	\item The generating curve of \textit{Tube 3} leads to a longer moving distance for infinitesimal elements of the cross section, and causes a larger energy input for resistance.
\end{itemize}

It is obvious that less energy consumed is needed for \textit{Tube 1} than \textit{Tube 2} and \textit{Tube 3}. By analyzing DOF under \emph{Definition \ref{def: dof}}, the same result as the intuition is obtained.

\subsection{Problem Formulation}

With the descriptions above, some assumptions are imposed to get the main problem of this paper.

\begin{itemize}
	\item \textbf{Assumption 1.} Let the density of the cross section $\mathcal{T}_s(\theta,\lambda)$ be its mass per unit length. Then the cross section has a uniform density and a constant mass as its shape changes during the process of passing through the virtual tube. 
	
	\item \textbf{Assumption 2.} The cross section moves along the generating curve at a constant speed $v_{\rm c}$ for any $s \in \mathcal{D}_s$. 
	
	\item \textbf{Assumption 3.} Each infinitesimal element of the cross section	receives a resistance whose direction is opposite its movement and magnitude proportional to the speed and mass.
	Denote the resistance coefficient by $\mu $, the mass and speed of the
	infinitesimal element as $\mathrm{d}m$ and $\bf{v}$ respectively, then the
	resistance ${\bf{f}}_{\rm r}$ is formulated as follows: 
	\begin{equation}
		\frac{\mathrm{d}{\bf{f}}_{\rm r}}{\mathrm{d}m}=-\mu \Vert \bf{v}\Vert \bf{v}.
		\label{equ: resistance}
	\end{equation}
\end{itemize}
\begin{remark}
	Eq.~(\ref{equ: resistance}) is derived from the well-established formula for calculating air resistance in aerodynamics, as outlined in reference \cite{airforce}. This formula indicates that resistance is directly proportional to the square of the speed and the resistance coefficient of the object in motion within the air.
\end{remark}
Under \textit{Assumptions 1-3} and \textit{Definitions \ref{def: 2d virtual tube}-\ref{def: dof}} , we will calculate and verify DOF for two-dimensional virtual tubes in the form of Eq.~(\ref{equ: 2d virtual tube}).

\section{A Measure of DOF and Modification}
\subsection{A DOF Measure}
\label{sec: dof modification}
\subsubsection{Mechanics of cross section infinitesimal elements}

DOF for a virtual tube is calculated based on the geometry of the tube and the mechanics of infinitesimal elements of the cross section. A cross section is composed of countless infinitesimal elements, each of them moving along its own trajectory according to mechanical principles.

At the beginning, the cross section is at point ${\bm{\gamma}}(s_{0})$. Each infinitesimal element corresponds to an initial proportion parameter, denoted by $\lambda_{\rm init}$. Then the initial position of an arbitrary infinitesimal element $P_0$ is
\begin{equation}
	{\bf{p}}(s_{0}, \theta, \lambda_{\rm init}) = \bm{\gamma} \left( s_{0} \right) + \lambda_{\rm init} r\left( s_{0},\theta \right) \mathbf{n} \left( s_{0} \right) \cos \theta ,
	\label{equ: initial position of a point}
\end{equation}
where $\lambda_{\rm init}\in \mathcal{D}_{\lambda_{\rm init} } = [0,1]$.

Under \textit{Assumption 1}, $\lambda$ of each infinitesimal element is always changing to maintain a uniform density, and the value of $\lambda$ is related to $s$, $\lambda_{\rm init}$ and $\theta$. When the cross section moves to point ${\bm{\gamma}}(s)$, denoted parameter $\lambda$ of $P_0$ by $\lambda \left( s,\theta, \lambda_{\rm init} \right)$, the following relationship holds as
\begin{equation}
	\frac{\left( 1-\lambda_{\rm init}\right) r\left( s_{0},\theta \right) }{r\left(
	s_{0}, 0 \right) + r\left( s_{0},\pi \right) } = \frac{\left( 1-\lambda (s, \theta, \lambda_{\rm init} ) \right)r(s,\theta )}{r\left( s,0\right) +r\left( s,\pi \right) }.
	\label{equ: proportional relationship}
\end{equation}
Here, $\lambda \left( s,\theta, \lambda_{\rm init} \right)$ is solved as
\begin{equation}
	\lambda (s,\theta, \lambda_{\rm init} )=1-\frac{(1-\lambda_{\rm init})r(s_{0},\theta
	)\left( r(s,\pi )+r(s,0) \right)}{r(s,\theta )\left( r(s_{0},0)+r(s_{0},\pi ) \right)},
	\label{equ: proportion}
\end{equation}
which is a function of $s$, $\theta $ and $\lambda_{\rm init}$. So, the position of $P_0$ is expressed as
\begin{equation}
	{\bf{p}}(s, \theta, \lambda_{\rm init} ) = \bm{\gamma} \left( s \right) +\lambda (s, \theta, \lambda_{\rm init})r\left(s,\theta \right) \mathbf{n}\left( s\right)
	\cos \theta,
	\label{equ: position of element}
\end{equation}
which is also a function of $s$, $\theta $ and $\lambda_{\rm init}$.

The velocity and acceleration of $P_0$ are as follows:
\begin{equation}
	\dot{{\bf{p}}}(s, \theta, \lambda_{\rm init} )=\frac{\mathrm{d}{\bf{p}}(s, \theta, \lambda_{\rm init})}{\mathrm{d}t},
	\label{equ: element velocity}
\end{equation}
\begin{equation}
	\ddot{{\bf{p}}}(s, \theta,\lambda_{\rm init}) = \frac{\mathrm{d}^{2}{\bf{p}}(s, \theta, \lambda_{\rm init})}{\mathrm{d}t^{2}}.
	\label{equ: element acceleration}
\end{equation}
Denote the density of the cross section at $\gamma (s)$ be $\rho (s)$. Then according to the principle of mass conservation, relationship holds
\begin{equation}
	\rho (s_{0})\cdot \left( r(s_{0},0)+r(s_{0},\pi ) \right)=\rho (s)\cdot \left( r(s,0) + r(s,\pi) \right).
	\label{equ: uniform density}
\end{equation}
Each infinitesimal of the cross section moves under an external force and the resistance. Let the infinitesimal of the cross section be $\mathrm{d}\left( \lambda \left(s, \theta, \lambda_{\rm init} \right) r \left(s, \theta \right)\right) $, the mass of the area element be $\mathrm{d}m=\rho (s) \mathrm{d} \left( \lambda \left(s, \theta, \lambda_{\rm init} \right) r \left( s,\theta \right) \right)$, the external force be ${\bf{f}} \left( s, \theta, \lambda_{\rm init} \right)$. Then, according to Newton's second law, the kinetic equation of the area element is
\begin{equation}
	\mathrm{d}m\cdot \ddot{{\bf{p}}} \left( s, \theta, \lambda_{\rm init} \right) = -\mu \Vert \dot{{\bf{p}}
	} \left(s, \theta, \lambda_{\rm init} \right) \Vert \dot{{\bf{p}}} \left( s, \theta, \lambda_{\rm init} \right){\mathrm{d}} m + {\bf{f}} \left( s, \theta, \lambda_{\rm init} \right),
	\label{equ: element kinetic quation}
\end{equation}
where $-\mu \Vert \dot{{\bf{p}}} \left( s, \theta, \lambda_{\rm init} \right) \Vert \dot{{\bf{p}}} \left( s, \theta, \lambda_{\rm init} \right) \mathrm{d}m$ is the resistance. Abbreviating ${\bf{p}} \left( s, \theta, \lambda_{\rm init} \right)$ to ${\bf{p}}$, the external force on the infinitesimal is
\begin{equation}
	\begin{aligned} {\bf{f}} \left( s, \theta, \lambda_{\rm init} \right) &= {\rm{d}} m \cdot \ddot{{\bf{p}} } + \mu \Vert \dot{{\bf{p}}} \Vert \dot{{\bf{p}}} {\rm{d}} m\\ 
	&= \left( \ddot{{\bf{p}} } + \mu \Vert \dot{{\bf{p}}}
	\Vert \dot{{\bf{p}}} \right) \cdot \rho \left( s \right) {\rm{d}} \left( \lambda \left( s, \theta, \lambda_{\rm init} \right) r
	\left( s, \theta \right) \right). 
	\end{aligned}
	\label{equ: external force}
\end{equation}

\subsubsection{Expression of DOF}

The work done by the external force to the infinitesimal during the whole process is 
\begin{equation}
	\begin{aligned} 
		E(\theta, \lambda_{\rm init}) &= \int_{{\bf{p}} (s_0, \theta, \lambda_{\rm init})}^{{\bf{p}} (s_f, \theta, \lambda_{\rm init})} {\bf{f}}(s, \theta, \lambda_{\rm init})^{\rm{T}} {\rm{d}} {{\bf{p}}} (s, \theta, \lambda_{\rm init})\\ &=
		\int_{s_0}^{s_f} {\bf{f}} (s, \theta, \lambda_{\rm init})^{\rm{T}} \frac{{\rm{d}} \left( {{\bf{p}}} (s, \theta, \lambda_{\rm init}) \right)}{{\rm{d}} s} {\rm{d}} s.
	\end{aligned}
	\label{equ: element work}
\end{equation}
The total external work done to the cross section is
\begin{equation}
	E=\int_{\lambda_{\rm init}}{\int_{\theta}E(\theta, \lambda_{\rm init} )\mathrm{d}\theta 
		\mathrm{d}\lambda_{\rm init}}.
	\label{equ: universal total work}
\end{equation}
For two-dimension virtual tubes, $\theta =0$ or $\pi$, Eq.~(\ref{equ: universal total work}) is expressed as
\begin{equation}
	E=\int_{0}^{1}E(0, \lambda_{\rm init}) \mathrm{d} \lambda_{\rm init} + \int_{0}^{1}E(\pi, \lambda_{\rm init}) \mathrm{d} \lambda_{\rm init}.
	\label{equ: 2d universal total work}
\end{equation}
With the average width obtained by Eq.~(\ref{equ: average width}), DOF is 
\begin{equation}
	DOF=\frac{\overline{l}}{E}.
	\label{equ: dof}
\end{equation}

\begin{figure}[!thbp]
	\centering
	\includegraphics[width=0.7\linewidth]{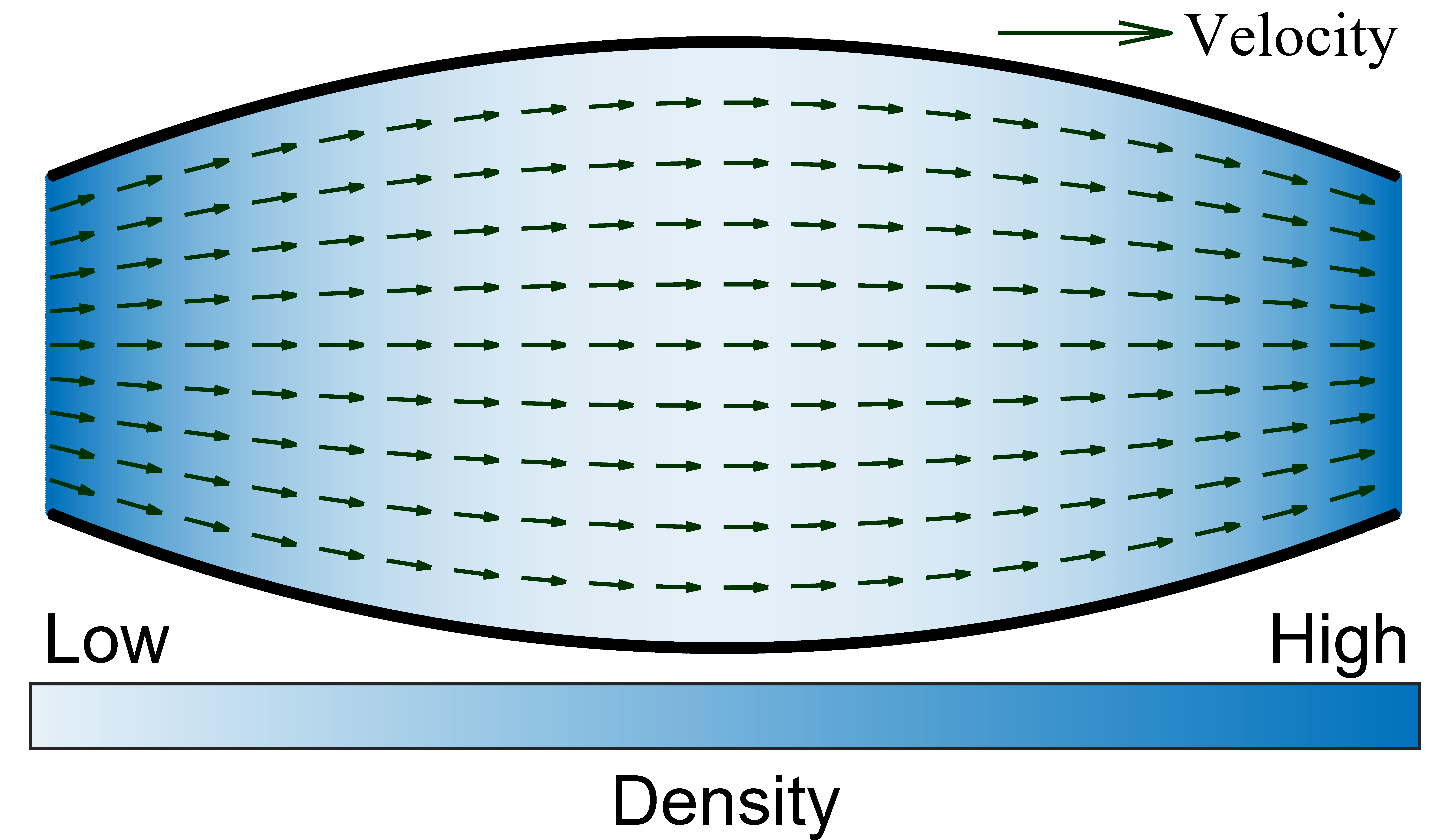}
	\caption{Velocity field and density field of a virtual tube}
	\label{fig: field}
\end{figure}

For $s \in \mathcal{D}_s$, $\lambda_{\rm init} \in \mathcal{D}_{\lambda_{\rm init}}$, $\theta \in \mathcal{D}_{\theta}$, velocity $\dot{{\bf{p}}} \left( s, \theta, \lambda_{\rm init} \right)$ and the density of cross section at different positions form a velocity vector field and a density field respectively, as in Fig.~\ref{fig: field}, which can be viewed as a \emph{virtual flow field} \footnote{The \emph{flow field} is described by several functions of space and time, which represent the variation of the flow dynamics properties, such as flow velocity, pressure, density and temperature \cite{flowfield}.}. This is different from the vector field which only considers velocity.

\subsection{Measure Analysis}
In \emph{Definition \ref{def: dof}}, the minimum input energy is required. In this section, we will show that the proposed measure (\ref{equ: dof}) satisfies this requirement. According to \textit{Assumption 1},\textit{\ }a cross section has a uniform density and a constant mass. A question arises whether the input energy will be decreased when infinitesimal elements change their order at two different cross sections. Before finding the answer, the trajectory of an infinitesimal element is defined.

\begin{Def}
	For an infinitesimal element whose initial position is expressed in ${\bf{p}}(s_{0}, \theta, \lambda_{\rm init})$, its trajectory in the virtual tube denoted by ${\mathcal{H}} \left( \theta, \lambda_{\rm init} \right)$ is
	\begin{equation}
		{\mathcal{H}}\left( \theta, \lambda_{\rm init} \right) =\bigcup\limits_{s\in \mathcal{D}_{s}}{\bf{p}}(s, \theta, \lambda_{\rm init}).
		\label{equ: trajectory def}
	\end{equation}
\end{Def}

\begin{figure}[tbp]
	\centering
	\subfigure[Example 1]{
		\centering
		\includegraphics[height=0.25\linewidth]{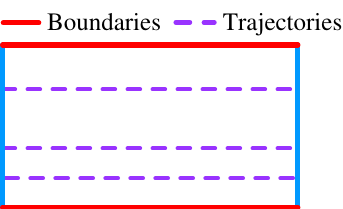}
		\label{subfig: rational example 1}
	}  
	\subfigure[Example 2]{
		\centering
		\includegraphics[height=0.25\linewidth]{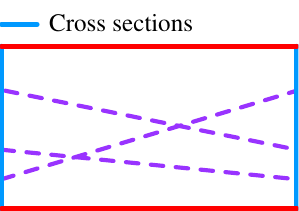}
		\label{subfig: rational example 2}
	}  
	\caption{Examples of different trajectories within the same virtual tube}
	\label{fig: rational example}
\end{figure}

For the same segment of a virtual tube, different trajectories of infinitesimal elements affects the amount of energy input. As in Fig.~\ref{fig: rational example}, there are intersecting trajectories in Fig.~\ref{subfig: rational example 2}. The total length of trajectories in Fig.~\ref{subfig: rational example 2} is longer than that in Fig.~\ref{subfig:
rational example 1}. The difference in trajectories affects the total energy
input to form a virtual tube. \textit{Definition \ref{def: dof}} implies that DOF is related to the minimum energy input. This subsection is to explain that the energy input calculated with the method stated in \emph{Section \ref{sec: dof modification}} is the minimum energy input.

To verify the proposed measure (\ref{equ: dof}), \emph{Proposition \ref{prop: no intersecting trajectories}} is proposed to state that the trajectories of different infinitesimal elements of the cross section do not intersect with each other.

\begin{prop}
	For any $s\in \mathcal{D}_{s}$, $ \lambda_{1}, \lambda_{2}\in \mathcal{D}_{\lambda_{\rm init}}$, $\theta_{1}, \theta_{2}\in \mathcal{D}_{\theta }$, if 
	\begin{equation}
		\theta_{1}=\theta_{2},\ \lambda_{1}\neq \lambda_{2}  \notag
	\end{equation}
	or 
	\begin{equation}
		\theta_{1}\neq \theta_{2},\ \vert \lambda_1 \vert + \vert \lambda_2 \vert \neq 0,  
		\notag
	\end{equation}
	then 
	\begin{equation}
		{{\bf{p}}}\left( s, \theta_{1}, \lambda_{1}\right) \neq {{\bf{p}}}\left(s, \theta_{2}, \lambda_{2}\right).
		\label{equ: no intersecting trajectories}
	\end{equation}
	Namely, 
	\begin{equation}
		{\mathcal{H}}\left( \theta_{1}, \lambda_{1}\right) \cup {\mathcal{H}} \left( \theta_{2}, \lambda_{2}\right) = \varnothing,
	\end{equation}
	for any $\lambda_1$, $\lambda_2$, $\theta_1$, $\theta_2$ that satisfy the conditions previously mentioned, which means there are no intersection between trajectories of different infinitesimal elements.
	\label{prop: no intersecting trajectories}
\end{prop}

\textbf{Proof}. See \emph{Appendix A}.

\begin{figure}[tbp]
	\centering
	\includegraphics[width=0.6\linewidth]{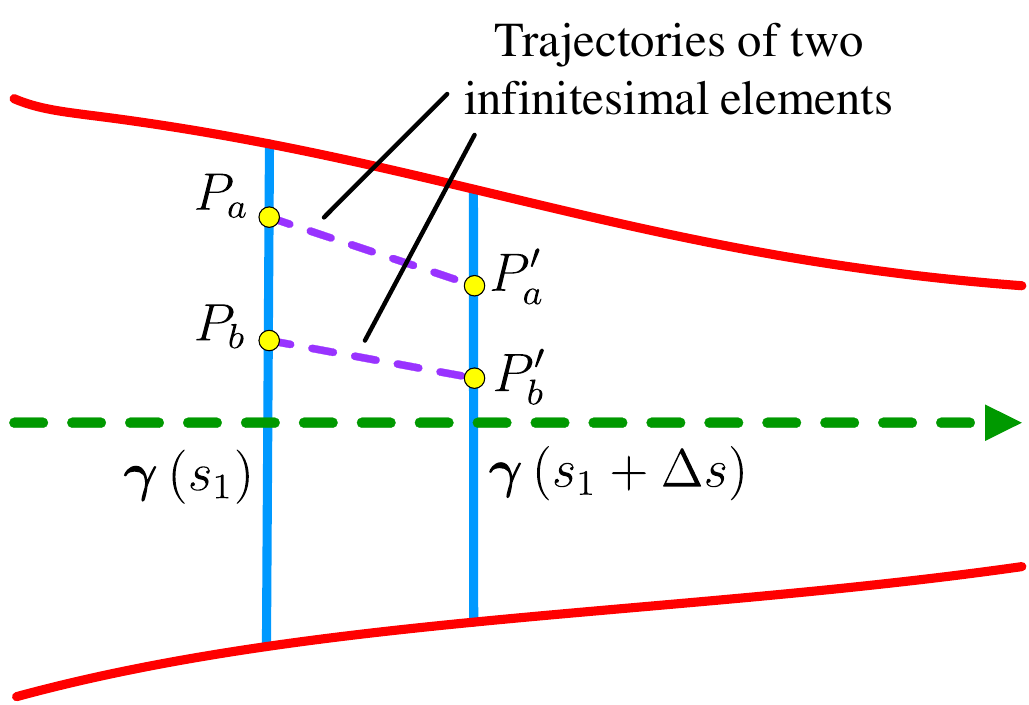}  
	\caption{Trajectories of infinitesimal elements in the cross section}
	\label{fig: 2d tube element}
\end{figure}

Next, we will show that trajectories without intersections cost the minimum energy input. The virtual tube can be divided into infinitesimal segments as in Fig.~\ref{fig: 2d tube element}. In each segment, the generating curve and boundaries of the virtual tube, as well as trajectories of infinitesimal elements are approximate to very short line segments. A cross section moves from ${\bm{\gamma}}\left( s_{1}\right)$ to ${\bm{\gamma}}\left( s_{1} + \Delta s\right)$, where $s_{1},s_{1}+\Delta s\in \mathcal{D}_{s}$. Consider two
given elements $P_a$, $P_{b}$ on cross section $\mathcal{T}_{s_{1}}\left( {\theta ,\lambda }\right) $, and the corresponding points $P_a^{\prime }$, $P_b^{\prime }$ on cross section $\mathcal{T}_{s_{1}+\Delta s}\left( {\theta ,\lambda }\right) $. The distance from $P_a$ to the upper boundary is shorter than that of $P_b$, and the same relationship exist between $P_a^{\prime }$ and $P_b^{\prime}$, which ensures that their positions do not coincide. For $\Delta s\rightarrow 0$ and the curvature of the generating curve is limited, we can presume that $P_a P_b$ and $P_a^{\prime }P_b^{\prime}$ are almost parallel, and elements $P_a$ and $P_b$ move in uniform motion. 

\begin{figure}[!t]
	\centering
	\includegraphics[width=0.8\linewidth]{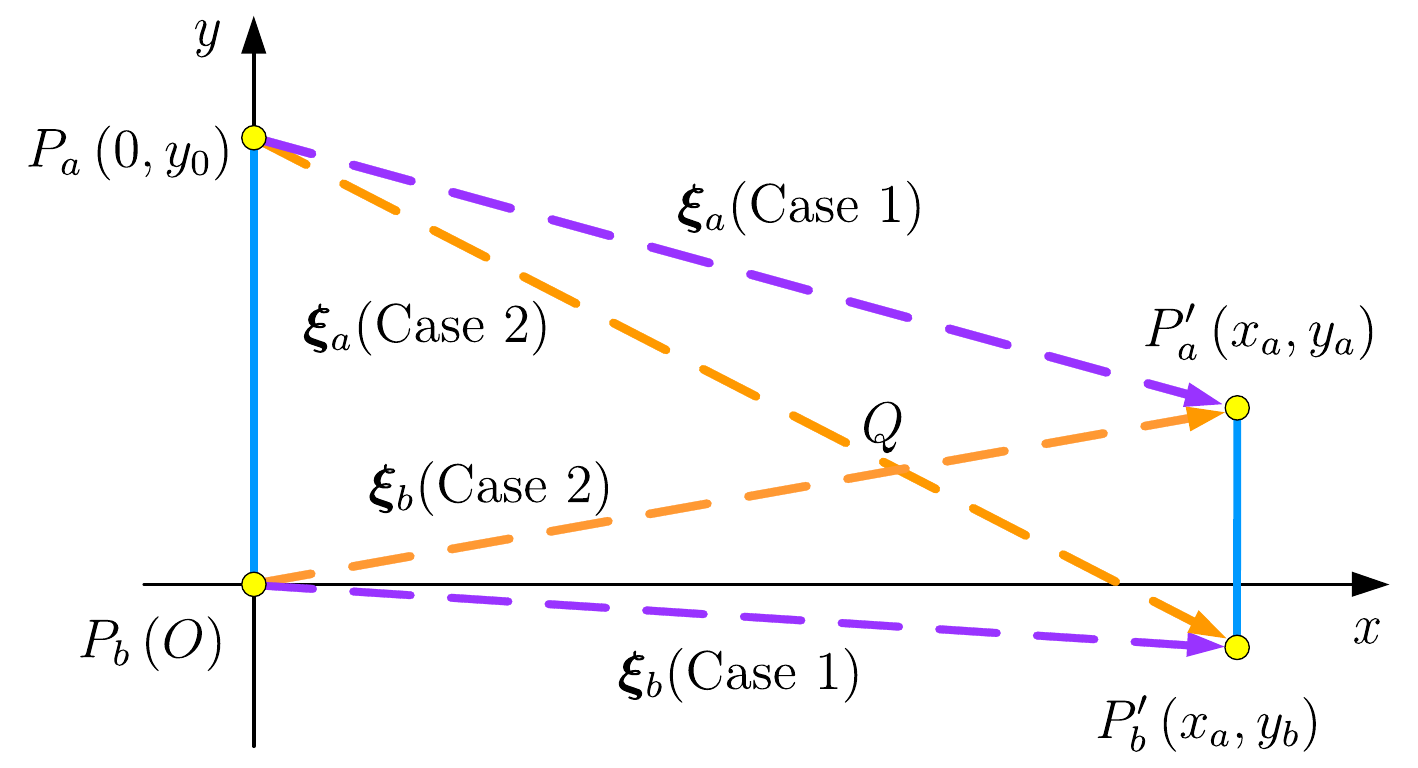}  
	\caption{Trapezoid $P_a P_b P_b^{\prime} P_a^{\prime}$ and two cases of position allocation}
	\label{fig: trapezoid example}
\end{figure}

To make further demonstration, trapezoid $P_a P_b P_b^{\prime}P_a^{\prime}$
is extracted alone, as in Fig.~\ref{fig: trapezoid example}. 
Two elements at $P_a$ and $P_b$ are to reach their destination at $P_a^{\prime}$ and $P_b^{\prime}$ in uniform motion. Denote the displacements of $P_a$ and $P_b$ as vectors ${\bm{\xi}}_a$ and ${\bm{\xi}}_b$ respectively. The cross section moves $\Vert \Delta {\bm{\gamma}} \left( s \right) \Vert$ at speed $v_{\rm c}$. Because of the properties of parallel lines, we can easily deduce that the speed of elements $P_a$ and $P_b$ are $v_a = \frac{\Vert {\bm{\xi}}_a \Vert}{\Vert \Delta {\bm{\gamma}} \left(s\right) \Vert} v_{\rm c}$ and $v_b = \frac{\Vert {\bm{\xi}}_b \Vert}{\Vert \Delta {\bm{\gamma}} \left(s\right) \Vert} v_{\rm c}$ respectively. There are two cases of position allocation.

\begin{itemize}
	\item \textbf{Case 1.} Element $P_a$ moves to $P_a^{\prime}$ while element $P_b$ moves to $P_b^{\prime}$. In this case their trajectories has no
	intersection, and ${\bm{\xi}}_a = \overrightarrow{P_a P_a^{\prime }}$, ${\bm{\xi}}_b = \overrightarrow{P_b P_b^{\prime }}$. 
	
	\item \textbf{Case 2.} Element $P_a$ moves to $P_b^{\prime}$ while element $P_b$ moves to $P_a^{\prime}$. Their trajectories intersect at point $Q$, and ${\bm{\xi}}_a = \overrightarrow{P_a P_b^{\prime }}$, ${\bm{\xi}}_b = \overrightarrow{P_b P_a^{\prime }}$.
\end{itemize}
Under \textit{Assumption 1}, the mass of elements $P_a$ and $P_b$ are
the same, set as $m_{\rm p}$ for convenience. According to Newton's Law and Eq.~(\ref{equ: resistance}), the external forces impacting $P_a$ and $P_b$ are 
\begin{equation}
	{\bf{f}}_{a}=\mu m_{\rm p} v_{a}^{2}\frac{{\bm{\xi}}_{a}}{\Vert {\bm{\xi}}_{a}\Vert },
	\notag
\end{equation}
\begin{equation}
	{\bf{f}}_{b}=\mu m_{\rm p}v_{b}^{2} \frac{{\bm{\xi}}_{b}}{\Vert {\bm{\xi}}_{b}\Vert}
	\notag
\end{equation}
respectively. Thus the total external work done to $P_a$ and $P_b$ is 
\begin{equation}
	\begin{aligned}
		E_{\rm p}
		&={\bf{f}}_{a}^{\mathrm{T}}{\bm{\xi}}_{a}+{\bf{f}}_{b}^{\mathrm{T}}{\bm{\xi}}_{b} \\
		& = \mu m_{\rm p} v_{a}^{2} \frac{ {\bm{\xi}}_{a}^{\mathrm{T}} {\bm{\xi}}_{a} }{ \Vert {\bm{\xi}}_{a} \Vert} + \mu m_{\rm p} v_{b}^{b} \frac{ {\bm{\xi}}_{b}^{\mathrm{T}} {\bm{\xi}}_{b} }{ \Vert {\bm{\xi}}_{b}\Vert } \\
		& =\frac{\mu m_{\rm p} v_{\rm c}^{2}}{\Vert \Delta {\bm{\gamma}}\left( s\right) \Vert^{2}}\cdot \left( \Vert {\bm{\xi}}_{a}\Vert ^{3}+\Vert {\bm{\xi}}_{b} \Vert^{3} \right) .
	\end{aligned}
	\label{equ: energy for case 1, 2}
\end{equation}

For $\frac{\mu m_{\rm p} v_{\rm c}^{2}}{\Vert \Delta {\bm{\gamma}}\left( s\right)\Vert ^{2}}>0$, to compare the energy input of two cases, $\Vert {\bm{\xi}}_{a}\Vert ^{3}+\Vert {\bm{\xi}}_{b}\Vert ^{3}$ is to be compared for \textit{Case 1} and \textit{Case 2}. \emph{Proposition \ref{prop: trapezoid}} about the relationship of the cube of legs and diagonals of a trapezoid is proposed.

\begin{prop}
	In trapezoid $P_a P_b P_b^{\prime } P_a^{\prime}$, where $P_a P_b$, $P_a^{\prime} P_b^{\prime}$ are the bases, there exists  
	\begin{equation}
		\vert P_a P_b^{\prime} \vert^3 + \vert P_b P_a^{\prime} \vert^3 > \vert P_a P_a^{\prime} \vert^3 + \vert P_b P_b^{\prime} \vert^3.
	\end{equation}
	\label{prop: trapezoid}
\end{prop}

\textbf{Proof}. See \emph{Appendix B}.

Thus, $\Vert {\bm{\xi}}_{a}\Vert ^{3}+\Vert {\bm{\xi}}_{b}\Vert ^{3}$ in \emph{Case 2} is greater than that in \textit{Case 1}. According to Eq.~(\ref{equ: energy for case 1, 2}), intersecting trajectories cost more external energy than non-intersecting ones. Intersections between trajectories of area
elements should be avoided to get the minimum energy input. According to \emph{Proposition \ref{prop: no intersecting trajectories}}, in the process of integration, the trajectories do not intersect, thus the result is the minimum energy input.
\begin{remark}
	\label{remark: explanation of the usage}
	The difficulty for a swarm to pass through a virtual tube, as measured by the DOF (Eqs.~(\ref{equ: initial position of a point}) to (\ref{equ: dof})), is dependent on the movement of the cross section. The proportional relationship in Eq.~(\ref{equ: proportional relationship}) is maintained so that the infinitesimal elements of the cross section retain their arrangement order and the energy input is minimized. In reality, the swarm moving in a virtual tube in different formations rarely remains within a cross section. The trajectories of the robots may intersect due to the use of various control algorithms.
\end{remark}

\subsection{Modified Measure}
The expressions of Equations~(\ref{equ: element velocity})-(\ref{equ: dof}) are based on the original shape of the virtual tube. The expansion of the whole swarm in an expanding part of the tube is not complete for it is time-consuming for the swarm to fill the expanding part. In consideration of the fact, the width of expanding virtual tubes should be modified by restriction of the expansion rate.

For virtual tubes whose $r \left( s, 0\right) = r \left(s, \pi \right)$, $l \left( s \right) = 2 r \left(s, \theta \right)$, where $\theta \in \left\{0, \pi \right\}$, radius $r\left( s, \theta \right) $ is modified to $
{r_{\rm m}}\left( s, \theta \right)$ to modify the width. Therefore, the modified virtual tube is denoted by
\begin{equation}
	{\mathcal{T}} \left( s, {r_{\rm m}}, \theta \right) = \bm{\gamma} \left( s \right) +
	\lambda {r_{\rm m}} \left( s, \theta \right) \mathbf{n} \left( s \right) \cos \theta.
	\label{equ: modified virtual tube}
\end{equation}

The density of the cross section shall increase when the tube shrinks but not decrease too much on the other hand.
We assume that the radius will expand to $\alpha r\left(s_{0}, \theta \right)$ at most, where $\alpha >1$ is the maximum expansion proportion, so that the minimum density is $\frac{1}{\alpha }\rho (s_{0})$. Taking robotic swarm dynamics into consideration, we presume that the free expansion speed for cross section is $\beta$, if the expansion is not restricted by the boundaries and the radius has not reach the maximum. Thus, the speed for expansion at $\left(s + \Delta s \right)$ is 
\begin{align}
	\nonumber
	&{\left. \frac{\mathrm{d} r_{\rm m} \left( s, \theta \right) }{\mathrm{d}s} \right|} _{s + \Delta s} = \\
	&\left\{
	\begin{aligned}
		& 0, && r_{\rm m} \left( s, \theta \right) \ge \alpha r \left( s_0, \theta \right) \\
		& \left. \frac{{\rm d} r \left(s, \theta \right)}{{\rm d} s} \right|_s, && r \left(s, \theta \right) \le r_{\rm m} \left( s, \theta \right) < \alpha r \left( s_0, \theta \right) {\rm and} \left. \frac{{\rm d} r \left(s, \theta \right)}{{\rm d} s} \right|_s < \beta \\
		& \beta, && else.
	\end{aligned}
	\right.
	\label{equ: expansion speed}
\end{align}

\begin{figure}[!t]
	\centering
	\includegraphics[width=0.65\linewidth]{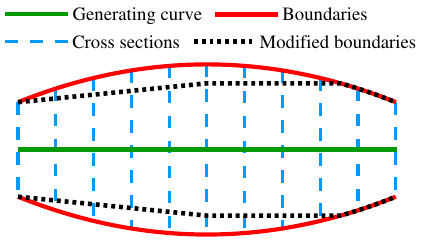}  
	\caption{An expanding virtual tube before and after the modification}
	\label{fig: modified expanding}
\end{figure}

The plot for a modified expanding virtual tube is shown in Fig.~\ref{fig: modified expanding}.

\section{Simulation}
To verify the methods proposed above, several kinds of typical two-dimensional virtual tubes are listed and numerical calculations and swarm control simulations are performed. Robot model and swarm control algorithm in study \cite{quan2023distributed} are applied to compare the results.

\subsection{Numerical Calculation Results}

Virtual tubes with the same starting cross section and finishing cross section are presented. The generating curve can be a straight line or a curve, the radius also changes differently.
\begin{figure*}[!bthp]
	\centering
	\includegraphics[width=1.0\columnwidth]{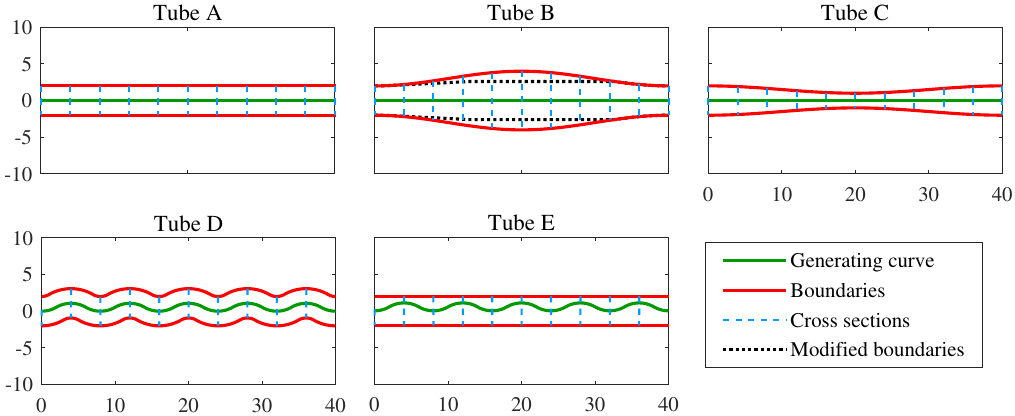}  
	\caption{Five typical virtual tubes}
	\label{fig: five virtual tubes}
\end{figure*}
The five tubes in Fig.~\ref{fig: five virtual tubes} have the same cross section of two parallel line segments of length 4 m, with a distance of 40 m. The speed of the moving cross section, denoted by $v_{\rm c}$, is set to 1 m/s, and total mass is set to 1 kg in order to calculate DOF in a standardized condition of a unit speed and a unit mass. The resistance coefficient is 0.5. Numerical calculation is made and the results are listed in Table \ref{tab: 2d calculation}.

\noindent\textbf{1. Virtual tubes with straight generating curves}

\emph{Tube A}-\emph{Tube C} have the same straight generating curve. Radii of different virtual tubes change differently and that causes the difference in DOF. Expanding \emph{Tube B} is modified with parameters $\alpha = 1.3$ and $\beta = 0.05$. As illustrated in Table~\ref{tab: 2d calculation}, \emph{Tube B} with an expanding radius exhibits the greatest DOF. The energy input of both the expanding \emph{Tube B} and the shrinking virtual \emph{Tube C} is greater than that of \emph{Tube A}. This is due to the fact that the elements' trajectories are curving, resulting in a longer total distance. However, the average width of \emph{Tube B} is greater than that of \emph{Tube A}, thus resulting in a larger DOF for \emph{Tube B}. This indicates that expanding virtual tubes have a greater DOF.

\noindent\textbf{2. Virtual tubes with curved generating curves}

\emph{Tube D} and \emph{Tube E} have generating curves that composed of several arcs. \emph{Tube D} has a constant radius as \emph{Tube A}, but in the curving \emph{Tube D}, the cross section has to move along a longer generating curve, and thus a larger energy input.

\emph{Tube E} has the same boundary as \emph{Tube A}, but with different generating curves. This means how the cross section passes through the virtual tube is different. In \emph{Tube E}, the curved generating curve and the constantly-changing cross section lead to more energy input than \emph{Tube A}.

\begin{table*}[!tbp]
	\caption{The calculation and simulation results of virtual tubes in  Fig.~\protect\ref{fig: five virtual tubes}}
	\label{tab: 2d calculation}
	\centering
	\scalebox{0.8}{
	\begin{tabular}{llccccc}
		\toprule
		\multicolumn{2}{l}{Virtual tube number}
		& A        & B        & C       & D        & E  \\
		\midrule
		\multicolumn{2}{l}{Minimum energy input (J)}
		&19.9950	&20.0143	&20.0259	&25.2251
		&24.2000 \\
		\midrule
		\multicolumn{2}{l}{blue}{Average width (m)}
		&4.0000 	&4.8454 	&3.0002 &4.0000 	&4.1964 \\
		\midrule
		\multicolumn{2}{l}{blue}{DOF (${\rm m} \cdot {\rm J}^{-1} \cdot 10^{-1} $)}
		&2.0005 	&2.4210		&1.4982 	&1.5857 	&1.7340 \\
		\midrule
		\multicolumn{2}{l}{Mean of Average crossing time (s)}
		&33.0909 	&29.6621 	&36.1146 	&33.8535 	&35.5416 \\ 
		\bottomrule
	\end{tabular}
	}	
\end{table*}

\subsection{Simulation Results}

The situation that one robotic swarm composed of 50 robots is considered. The controller developed in \cite{quan2023distributed} is capable of guiding a swarm through a virtual tube with two known borders and a generating curve. The controller incorporates three control terms. The line associated with the first term directs the robots to move primarily in the direction of the generating curve, ultimately reaching the finishing line from the start. The second term, which addresses robot avoidance, ensures that the robots avoid collisions. The third term, which pertains to maintaining the tube, guarantees that all robots remain within the virtual tube throughout the process.

To ensure that all robots in the swarm remain within the virtual tube at the outset of the simulation, a smooth buffer tube is incorporated at the front of the virtual tubes under consideration but is excluded from the calculation of DOF. The buffer tube is 25 m in length, with a 16 m starting line and a 4 m finishing line.
\begin{figure}[!htbp]
	\centering
	\includegraphics[width=1.0\linewidth]{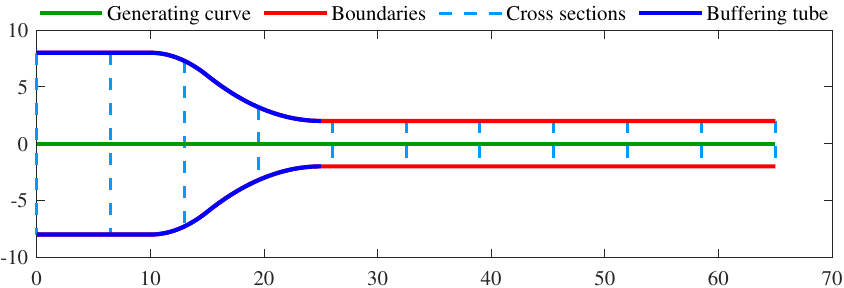}
	\caption{Virtual Tube A with a bufffering tube}
	\label{fig: funnel tube}
\end{figure}

The robots are randomly arranged in a rectangular area at the beginning, which is defined by $\left\{(x,y) | \left(0 \le x \le 10, -7.5 \le y \le 7.5\right)\right\}$. The robots in the swarm has a different random speed distribution according to their positions. The robot in the front half of the area has speed that obeys uniform distribution $U(1,1.25) (\rm m/s)$, while the robot in the back half of the area has speed that obeys uniform distribution $U(1.75,2) (\rm m/s)$. The robots situated towards the rear of the swarm are observed to exhibit superior velocity compared to those positioned at the front. This phenomenon results in the robots at the rear of the swarm overtaking those at the front. Consequently, the swarm assumes a variable shape and displays a propensity to occupy the virtual tube. Additionally, it is conceivable that the rapid robots may surpass the slower ones in terms of velocity.
\begin{figure}[!thbp]
	\centering
	\includegraphics[width=1.0\linewidth]{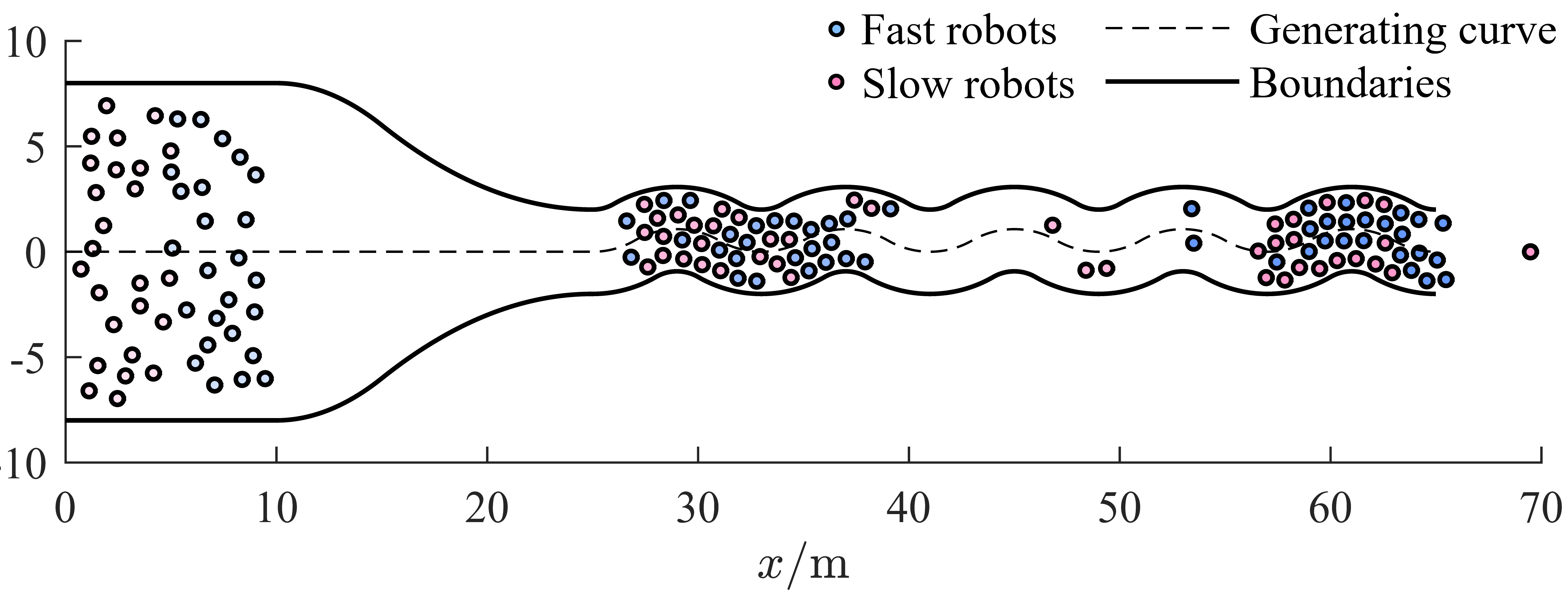}
	\caption{The snapshots illustrate the initial and intermediate stages of a simulation. Robots operating at a slower pace are colored blue, while those moving at a faster velocity are painted pink. During the collective movement of the swarm, some of the faster robots are observed to surpass the slower ones at the front.}
	\label{fig: simulation pictures}
\end{figure}

\emph{Average crossing time} is chosen to compare the difficulty for the swarm to pass through the virtual tube. It is expressed as follows.
\begin{equation}
	\overline{t} = \frac{1}{N} \sum_{i=1}^{N}t_i,
	\label{equ: average crossing time}
\end{equation}
where $N$ is the total number of the robots, $t_i$ is the time for the $i$th robot to pass from the end of the buffering tube to the end of the whole tube. A longer average crossing time results in greater challenges for the swarm to successfully navigate the environment.

The safety radius is set to $s_{\rm f} = 0.3 \ \rm m$ and the avoidance radius is set to $s_{\rm a} = 0.6 \ \rm m$. Simulations are carried out 10 times for each virtual tube in Fig.~\ref{fig: five virtual tubes}, and the average crossing time calculated. The results are given in Table \ref{tab: 2d calculation}.

In the case of virtual tubes with straight generating curves, the tube with a smaller DOF exhibits a longer average crossing time. This observation lends support to the notion that the method is effective for cases of straight-generating curves. The incremental DOF and decreasing average crossing time observed in \emph{Tubes B, A}, and \emph{C} indicate that a larger width can mitigate the difficulty encountered by the robotic swarm when passing through a straight virtual tube. Furthermore, the comparison between \emph{Tube A} and \emph{Tube E} illustrates that the shape of the generating curve affects the difficulty of the swarm traversing the virtual tube. This is because the generating curve is an input for the swarm control algorithm. From the trajectory plot of five arbitrary robots, it can be observed that the trajectories in \emph{Tube E} exhibit greater Z-turns and greater flexuosity than those in \emph{Tube A}. This results in a longer average crossing time. In general, \emph{Tubes A} and \emph{B} exhibit superior performance compared to \emph{Tubes C, D}, and \emph{E}. This suggests that a shrinking radius and a curved generating curve can significantly impact the quality of virtual tubes and that such characteristics should be avoided in practice whenever possible.
\begin{figure}[!ht]
	\centering
	\subfigure[Trajectories of robots in \emph{Tube A}]{
		\centering
		\includegraphics[width=0.85\linewidth]{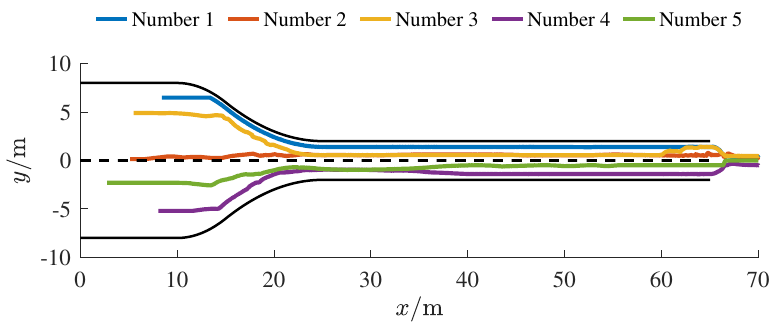}
		\label{subfig: straight trajectories}
	}  
	\subfigure[Trajectories of robots in \emph{Tube E}]{
		\centering
		\includegraphics[width=0.85\linewidth]{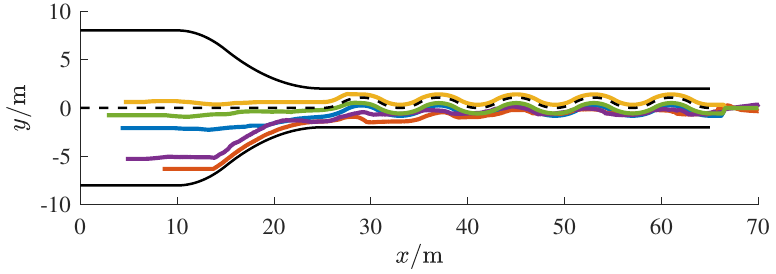}
		\label{subfig: curve trajectories}
	}  
	\caption{Trajectories of robots in virtual tube with different generating curves}
	\label{fig: trajectories}
\end{figure}

\begin{figure}[!tbp]
	\centering
	\includegraphics[width=0.95\linewidth]{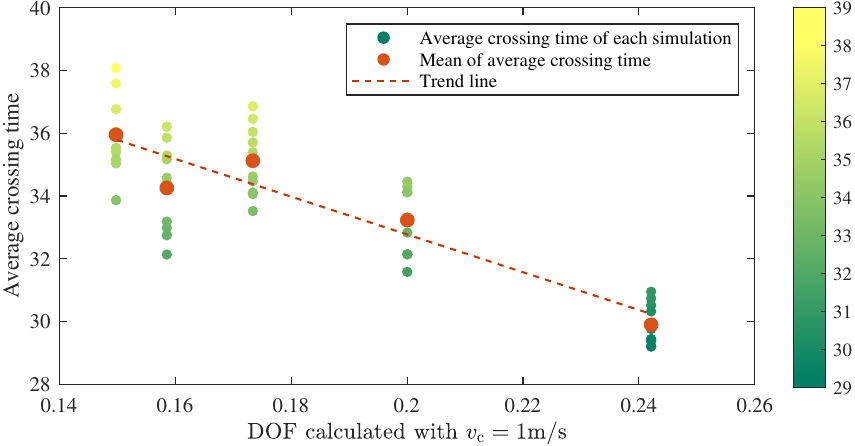}
	\caption{Scatter plot of the calculated DOF to the standardized average score}
	\label{fig: scatter dof}
\end{figure}

To investigate the correlation between the calculated DOF and the overall performance of the five tubes, a scatter plot is constructed, as illustrated in Fig.~\ref{fig: scatter dof}. The plot reveals a negative correlation between the average crossing time of each virtual tube and DOF, indicating that DOF can be utilized as a reference for the quality of the virtual tube.
\begin{remark}
	In Table \ref{tab: 2d calculation} and Fig.~\ref{fig: scatter dof}, the negative relationship between DOF and the standardized average score is not strict. The simulation results for \emph{Tube D} and \emph{Tube E} do not correspond with their calculated DOF. This is because the average crossing time is closely related to the controller applied. The controller in \cite{quan2023distributed} has a line approaching term which makes the robots have a velocity component in the direction of the generating curve. In view of the curved generating curve, cross sections \emph{Tube D} has a constant length and cross sections of \emph{Tube E} has a varying length. The robots moves slower because of the varying cross section. In practice, the robots guided by the controller do not move in the way of the cross section of the virtual tube, so the specific relationship is hard to obtain. The outcome may vary depending on the controller employed.
\end{remark}

\section{Conclusion}
In this paper, the definitions of the two-dimensional virtual tube and its DOF is presented. DOF is defined as the product of the average length of the virtual tube and the reciprocal of the minimum energy input for the cross section to move perpendicularly along the generating curve and form the virtual tube. A measure of DOF is derived and shown feasible to calculate the minimum energy input. Simulations of different two-dimensional virtual tubes shows that the radius and generating curve both affect its DOF. Straight generating curve and a larger width of a virtual tube makes a relatively high DOF, and it will be easier for the robotic swarm to pass through the virtual tube. In the future, the definitions and measures of DOF for more situations should be considered, and three-dimensional virtual tubes will be also deserved to take into consideration.

\appendix
\section{Proof of \emph{Proposition \ref{prop: no intersecting trajectories}}}

There are two conditions of (\ref{equ: no intersecting
	trajectories}):

\begin{itemize}
	\item $\theta_1 = \theta_2 = \theta$, $\lambda_1 \neq \lambda_2$. 
	
	\item $\theta_1 \neq \theta_2$ and $\vert \lambda_1 \vert + \vert \lambda_2 \vert \neq 0$. Without
	loss of generality, let $\theta_1 = 0$, $\theta_2 = \pi$.
\end{itemize}

(i) If $\theta_1 = \theta_2 = \theta$, from Eq.~(\ref{equ: proportion}), we obtain 
\begin{equation}
	\frac{\partial \lambda(s, \theta, \lambda_{\rm init})}{ \partial \lambda_{\rm init}} = \frac{r(s_0, \theta) \left( r (s, \pi) + r(s, 0) \right)}{r(s, \theta) \left( r(s_0,0) + r(s_0, \pi) \right)}.
	\notag
\end{equation}
For any $s \in \mathcal{D}_s$, $\theta \in \mathcal{D}_{\theta}$, there
exists 
\begin{equation}
	r \left(s, \theta \right) > 0.  \notag
\end{equation}
Then 
\begin{equation}
	\frac{\partial \lambda(s, \theta, \lambda_{\rm init})}{ \partial \lambda_{\rm init}} > 0, 
	\notag
\end{equation}
which means $\lambda(s, \theta, \lambda_{\rm init})$ is a monotonically increasing
function of $\lambda_{\rm init}$. For any $\lambda_1 \neq \lambda_2$, there is 
\begin{equation}
	\lambda(s, \theta, \lambda_{1}) \neq \lambda(s, \theta, \lambda_{2}),
	\notag
\end{equation}
that is, 
\begin{equation}
	\lambda(s, \theta_1, \lambda_{1}) \neq \lambda(s, \theta_2, \lambda_{2}).
	\notag
\end{equation}

(ii) If $\theta_1 = 0$, $\theta_2 = \pi$, we use proof by contradiction to show \emph{Proposition} \ref{prop: no intersecting trajectories}. Assume
that inequality (\ref{equ: no intersecting trajectories}) is wrong, and there exist $\lambda_1, \lambda_2 \in \mathcal{D}_{\lambda_{\rm init}}$, $\vert \lambda_1 \vert + \vert \lambda_2 \vert \neq 0$, s.t. 
\begin{equation}
	{{\bf{p}}} \left( s, 0, \lambda_1\right) = {{\bf{p}}} \left( s, \pi, \lambda_2 \right).
	\label{equ: if intersects}
\end{equation}
Substituting Eq.~(\ref{equ: position of element}) into (\ref{equ: if intersects}), we get 
\begin{equation}
	\lambda(s, 0, \lambda_1) \cdot r \left( s, 0 \right) + \lambda(s, \pi, \lambda_2) \cdot r \left( s, \pi \right) = 0.  \notag
\end{equation}
Substituting Eq.~(\ref{equ: proportion}) into the equation above, and
simplifying the formulation, we obtain 
\begin{equation}
	\frac{\left( r(s,0) + r(s, \pi) \right) \left( \lambda_1 r(s_0,0) + \lambda_2 r(s_0, \pi) \right)}{r(s_0,0) + r(s_0,\pi)} = 0.
	\label{equ: if position equal}
\end{equation}
For any $s \in \mathcal{D}_s$, $\theta \in \mathcal{D}_{\theta}$, there
exists 
\begin{equation}
	r \left(s, \theta \right) > 0.  \notag
\end{equation}
So, Eq.~(\ref{equ: if position equal}) is further simplified to 
\begin{equation}
	\lambda_1 r(s_0,0) + \lambda_2 r(s_0, \pi) = 0.  \notag
\end{equation}
Since $r(s_0,0) > 0$, $r(s_0, \pi) > 0$, there is 
\begin{equation}
	\lambda_1 = -\frac{r(s_0, \pi)}{r(s_0, 0)} \lambda_2.
	\label{equ: if lambda_1<0}
\end{equation}
For $\lambda_1, \lambda_2 \in \mathcal{D}_{\lambda_{\rm init}}$, if and only if $\lambda_1 = \lambda_2 = 0$ is Eq.~(\ref{equ: if lambda_1<0}) satisfied,
which contradicts $\vert \lambda_1 \vert + \vert \lambda_2 \vert \neq 0$. So Eq.~(\ref{equ: if intersects}) does not hold true. Thus, 
\begin{equation}
	{{\bf{p}}} \left( s, \theta_1, \lambda_1 \right) \neq {{\bf{p}}} \left( s, \theta_2, \lambda_2 \right).
	\notag
\end{equation}

\section{Proof of \emph{Proposition} \protect\ref{prop: trapezoid}}

Trapezoid $P_a P_b P_b^{\prime} P_a^{\prime}$ is placed in a Cartesian coordinate system in Fig.~\ref{fig: trapezoid example}. Vertex $P_a$ is arranged on Origin $O$, $P_a P_b$ is on the $y$ axis. The coordinates of points $P_a$, $P_a^{\prime}$, $P_b^{\prime}$ are shown in Fig.~\ref{fig: trapezoid example}. Without loss of generality, we
assume that $y_0 > 0$, $x_a > 0$, $y_a > y_b$.

Based on knowledge of plane analytic geometry, there are 
\begin{equation}
	\begin{aligned} \vert P_a P_a'\vert &= \sqrt{x_a^2 + \left( y_a - y_0 \right)^2},\\
	\vert P_b P_b'\vert &= \sqrt{x_a^2 + y_b^2},\\
	\vert P_a P_b'\vert &= \sqrt{x_a^2 + \left( y_b - y_0 \right)^2},\\
	\vert P_b P_a'\vert &= \sqrt{x_a^2 + y_b^2}.
	\end{aligned}
\end{equation}

Construct a function with respect to $y$ as
\begin{equation}
	g \left( y \right) = \left( \sqrt{x_a^2 + y^2} \right)^3 - \left( \sqrt{x_a^2 + \left( y - y_0 \right)^2} \right)^3.
	\label{equ: g(y)}
\end{equation}
The derivative of $g\left( y \right)$ is 
\begin{equation}
	g^{\prime }\left( y \right) = 3y \sqrt{x_a^2 + y^2} - 3\left( y - y_0	\right) \sqrt{x_a^2 + \left( y - y_0 \right)^2}.
	\label{equ: g'(y)}
\end{equation}

Construct another function with respect to $y$ as 
\begin{equation}
	h \left( y \right) = 3y \sqrt{x_a^2 + y^2}.  \label{equ: h(y)}
\end{equation}
The derivation of $h \left( y \right)$ is 
\begin{equation}
	h^{\prime}\left( y \right) = 3 \sqrt{x_a^2 + y^2} + \frac{3y^2}{\sqrt{x_a^2	+ y^2}}.
	\notag
\end{equation}

For any $y \in \mathbb{R}$, $x_a > 0$, there is $h^{\prime }\left( y \right) > 0$, so $h \left( y \right)$ is a strictly monotone increasing function with respect to $y$. As $y_0 > 0$, there is $y > y - y_0$. Then, 
\begin{equation}
	h \left( y \right) > h \left( y - y_0 \right).  \label{equ: h(y)>h(y-y0)}
\end{equation}

Compare Expressions (\ref{equ: g'(y)}), (\ref{equ: h(y)}) and (\ref{equ:
	h(y)>h(y-y0)}), and we get 
\begin{equation}
	g^{\prime }\left( y \right) = h \left( y \right) - h \left( y - y_0 \right)
	> 0.  \notag
\end{equation}

Thus, $g \left( y \right)$ is a strictly monotone increasing function with respect to $y$. For any $y_a > y_b$, there is 
\begin{equation}
	g \left( y_a \right) > g \left( y_b \right).
	\notag
\end{equation}
Substituting Eq.(\ref{equ: g(y)}) into the inequation above, we get 
\begin{equation}
	\left( \sqrt{x_a^2 + y_a^2} \right)^3 - \left( \sqrt{x_a^2 + \left( y_a -	y_0 \right)^2} \right)^3 > 
	\left( \sqrt{x_a^2 + y_b^2} \right)^3 - \left( \sqrt{x_a^2 + \left( y_b - y_0 \right)^2} \right)^3.
	\notag
\end{equation}
With some simple adjustment, we can finally get 
\begin{equation}
		\left( \sqrt{x_a^2 + \left( y_b - y_0 \right)^2} \right)^3 + \left( \sqrt{x_a^2 + y_a^2} \right)^3 >
		\left( \sqrt{x_a^2 + \left( y_a - y_0 \right)^2} \right)^3 + \left( \sqrt{x_a^2 + y_b^2} \right)^3,
	\notag
\end{equation}
which means 
\begin{equation}
	\vert P_a P_b ^{\prime} \vert^3+ \vert P_b P_a^{\prime}\vert^3 > \vert P_a P_a^{\prime}\vert^3 + \vert P_b P_b^{\prime} \vert^3.
	\nonumber
\end{equation}

\bibliographystyle{elsarticle-num} 
\bibliography{dof}

\end{document}